\documentclass[twocolumn]{article}

\usepackage[left=1in,right=1in,top=1in,bottom=1in]{geometry}
\usepackage{tikz}
\usetikzlibrary{calc,patterns,angles,quotes}

\usepackage{algorithm}
\usepackage[noend]{algpseudocode}
\usepackage{amsmath}
\usepackage{authblk}
\usepackage{url}
\usepackage{multirow}
\usepackage{wasysym}

\usepackage[explicit]{titlesec}
\renewcommand{\thesection}{\Roman{section}}
\renewcommand{\thesubsection}{\thesection.\Alph{subsection}}
\renewcommand{\thesubsubsection}{\thesubsection.\arabic{subsubsection}}

\titleformat{\section}{\large\scshape\centering}{\thesection.\space}{0pt}{#1}[]
\titlespacing*{\section}{0pt}{0.5\baselineskip}{0pt}
\titleformat{\subsection}{\normalsize\itshape}{\Alph{subsection}.\space}{0pt}{#1}[]
\titlespacing*{\subsection}{0pt}{0.5\baselineskip}{0pt}
\titleformat{\subsubsection}{\normalsize\itshape}{\arabic{subsubsection}.\space}{0pt}{#1}[]
\titlespacing*{\subsubsection}{0pt}{0.5\baselineskip}{0pt}
\usepackage{geometry}
\newgeometry{left=2cm, right=2cm, top=2.5cm,bottom=3.5cm}
\makeatletter
\renewcommand{\fnum@figure}{Fig. \thefigure}
\renewcommand{\fnum@table}{Tab. \thetable}
\makeatother

\let\oldbibitem\bibitem
\def\bibitem{\ifnum\value{enumiv}>0 \vspace{-0.7175\baselineskip} \fi\oldbibitem}

\usepackage{nopageno} 
\usepackage{geometry}

\newgeometry{left=2cm, right=2cm, top=2.5cm,bottom=3.5cm}

\renewcommand{\abstractname}{}

\title{\vspace{-1.0cm}\textbf{\normalsize Closing the gap: Optimizing Guidance and Control Networks through Neural ODEs}}

\author{Sebastien Origer} 
\author{Dario Izzo} 
\affil[1]{\emph{Advanced Concepts Team, European Space Research and Technology Centre (ESTEC), Noordwijk, The Netherlands.}}

\date{}  
\begin{document}

\maketitle

\begin{abstract}
\vspace{-1.15\baselineskip}
\textbf{\emph{\quad Abstract} - 
We improve the accuracy of Guidance \& Control Networks (G\&CNETs), trained to represent the optimal control policies of a time-optimal transfer and a mass-optimal landing, respectively.
In both cases we leverage the dynamics of the spacecraft, described by Ordinary Differential Equations which incorporate a neural network on their right-hand side (Neural ODEs). 
Since the neural dynamics is differentiable, the ODEs sensitivities to the network parameters can be computed using the variational equations, thereby allowing to update the G\&CNET parameters based on the observed dynamics.
We start with a straightforward regression task, training the G\&CNETs on datasets of optimal trajectories using behavioural cloning.
These networks are then refined using the Neural ODE sensitivities by minimizing the error between the final states and the target states.
We demonstrate that for the orbital transfer, the final error to the target can be reduced by $99\%$ on a single trajectory and by $70\%$ on a batch of $500$ trajectories. For the landing problem the reduction in error is around $98-99\%$ (position) and $40-44\%$ (velocity).
This step significantly enhances the accuracy of G\&CNETs, which instills greater confidence in their reliability for operational use.
We also compare our results to the popular Dataset Aggregation method (DaGGER) and allude to the strengths and weaknesses of both methods.}
\end{abstract}

\section{Introduction}
Guidance and Control Networks (G\&CNETs) represent an emerging technique that holds promise for on-board autonomy and the seamless integration of optimality principles into spacecraft and space agents \cite{sanchez2018real, izzo2021real, li2019neural, federici2021deep, hovell2020deep, cheng2018real, dario_seb_gcnet}. They serve as an alternative to model predictive control schemes (MPC) \cite{eren2017model}, capitalizing on the numerous improvements and advances arising from neural network-based research. 
Both Reinforcement Learning (RL) and Behavioural Cloning (BC) have already demonstrated successful implementations in training G\&CNETs for both space and drone-related tasks \cite{izzo2023optimality}. 
However no matter which learning paradigm one chooses, residual approximation errors lead to orbit injection errors that must be corrected at the cost of extra on-board propellant.
Therefore, after training a G\&CNET, it is important to further reduce the final mismatch between the targeted final orbit injection conditions and the ones achieved by the on-board neural guidance and control. 
This paper considers neural models to represent the optimal control policy for a time-optimal interplanetary transfer targeting a generic low-thrust Earth rendezvous starting from the asteroid belt \cite{dario_seb_gcnet} and a mass-optimal landing on the asteroid Psyche. 
We chose these two optimal control problems such that our study covers different timescales and problems of varying difficulty. 
The transfer is a complex low-thrust problem which lasts years. 
In contrast, the landing problem requires the G\&CNET to learn a discontinuous function representing a bang-bang control profile and lasts only minutes.
In our work we use the term Neural ODEs, popularized in \cite{chen2018neural}, to describe Ordinary Differential Equations which have an artificial neural network on their right-hand side.
We exploit the fact that, for fixed initial conditions, the solution to such a system depends only on the network parameters. 
We thus proceed to study the use of Ordinary Differential Equations (ODEs) sensitivities to the network parameters.
Since our Neural ODEs are differentiable, the variational equations, or equivalently Pontryagin's adjoint method, enable us to compute efficiently thousands of ODE sensitivities (state transition matrix).
These partial derivatives are used to inform a local search into the highly dimensional network parameters space, aligning with the recent trend of Neural ODEs. We use a simple gradient descent algorithm to update the G\&CNET parameters such as to minimize the mismatch between the final states and the target states.
\section{Methods}
\subsection{Optimal control problems}
\subsubsection*{Time-optimal interplanetary transfer}
\begin{figure*}[tb]
   \centering
\includegraphics[width=0.7\textwidth]{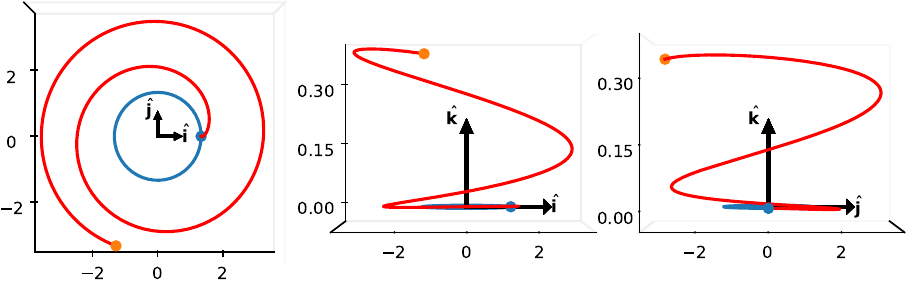}
\caption{Interplanetary transfer shown in rotating frame $\mathcal F$. Axis unit is AU.}\label{fig:nominal_transfer}
\end{figure*}

We consider the same time optimal, constant acceleration rendezvous with a body in a perfectly circular orbit of radius $R$ as in \cite{dario_seb_gcnet}.
Let $\mathcal F = \left[\hat{\mathbf i}, \hat{\mathbf j}, \hat{\mathbf k}\right]$ be a rotating frame with angular velocity $\boldsymbol \Omega= \sqrt{\frac{\mu}{R^3}} \hat {\mathbf k}$. In this way, the target body position $R\hat{\mathbf i}$ remains stationary in $\mathcal F$. The dynamics is described by the following ordinary differential equations:
\begin{equation}
\label{eq:dyn_transfer}
\left\{ 
\begin{array}{l}
    \dot{x} = v_x  \\
    \dot{y} = v_y \\
    \dot{z} = v_z \\
    \dot{v}_x = -\frac{\mu}{r^3}x + 2 \Omega v_y + \Omega^2 x +\Gamma i_x \\
    \dot{v}_y = -\frac{\mu}{r^3}y - 2 \Omega v_x + \Omega^2 y +\Gamma i_y \\
    \dot{v}_z = -\frac{\mu}{r^3}z +\Gamma i_z
\end{array}
\right.
\end{equation}
The state vector $\mathbf{x}_T$ (subscript $\Box_T$ to indicate the "transfer" optimal control problem) contains the position $\mathbf{r}=[x,y,z]$ and velocity $\mathbf{v}=[v_x,v_y,v_z]$ which are both defined in the rotating frame $\mathcal F$. Note that $r = \sqrt{x^2+y^2+z^2}$ and $\mu$ is the gravitational constant of the Sun.
The system is controlled by the thrust direction described by the unit vector $\mathbf{\hat{t}} = [i_x, i_y, i_z]$, generating an acceleration of magnitude $\Gamma$. The optimal control problem boils down to finding the optimal time-of-flight $t_f$ and a (piece-wise continuous) function for $\mathbf{\hat{t}}(t)$, where $t\in [t_0,t_f]$, such that, under the dynamics described by Eq.\ref{eq:dyn_transfer}, the state is steered from any initial state $\mathbf{r}_0$, $\mathbf{v}_0$ to the desired target state $\mathbf{r}_t=R\mathbf{\hat{i}}$, $\mathbf{v}_t=\mathbf{0}$. We are thus minimizing the following cost function:
$J = t_f-t_0 = \int^{t_f}_{t_0}  dt$ \cite{dario_seb_gcnet}. Let's solve this problem using Pontryagin's Maximum Principle \cite{pontryagin}, taking into account some useful tips from \cite{jiang2012practical}. Let $\mathcal{H}$ be the Hamiltonian:
\begin{equation}
\label{eq:hamiltonian}
\begin{split}
    \mathcal H(\mathbf{r},\mathbf{v},\boldsymbol{\lambda}_r,\boldsymbol{\lambda}_{\mathbf{v}},\mathbf{\hat{t}})= \boldsymbol{\lambda}_\mathbf{r}\cdot \mathbf{v}+ \\ \boldsymbol{\lambda}_\mathbf{v}\cdot\bigg(-\frac{\mu}{r^3}\mathbf{r}-2\boldsymbol \omega\times\mathbf v-\boldsymbol \omega\times\boldsymbol \omega\times\mathbf r+\Gamma\mathbf{\hat{t}}\bigg)+\lambda_J  
\end{split}
\end{equation}
where $\boldsymbol{\lambda}_\mathbf{r}$ and $\boldsymbol{\lambda}_\mathbf{v}$ are the co-sates functions and $\lambda_J$ is an additional constant coefficient used to multiple our cost function $J = \lambda_J(t_f-t_0)$. This additional constant increases numerical stability and offers an additional degree of freedom when performing the Backward Generation of Optimal Examples (BGOE) in Sec.\ref{subsec:BC} \cite{dario_seb_gcnet}. For a trajectory to be optimal the classical necessary condition tells us that thrust direction $\mathbf{\hat{t}}^*$ needs to minimize the Hamiltonian, hence:
\begin{equation}\label{eq:neccondition}
    \mathbf{\hat{t}}^*=-\frac{ \boldsymbol{\lambda}_{\mathbf v}}{|\boldsymbol{\lambda}_{\mathbf v}|}
\end{equation}
The augmented system of equations is then obtained by taking the derivatives of the Hamiltonian with respect to $\dot {\mathbf x}_T = \frac{\partial \mathcal H}{\partial \boldsymbol \lambda}$ and $\dot {\boldsymbol \lambda} = - \frac{\partial \mathcal H}{\partial \mathbf x_T}$:
\begin{equation}
\label{eq:augdyn}
\left\{ 
\begin{array}{l}
    \dot{\mathbf{r}} = \mathbf{v} \\
    \dot{\mathbf{v}} =  -\frac{\mu}{r^3}\mathbf{r}-2\boldsymbol \omega\times\mathbf v-\boldsymbol \omega\times\boldsymbol \omega\times\mathbf r-\Gamma\frac{ \boldsymbol{\lambda}_{\mathbf v}}{|\boldsymbol{\lambda}_{\mathbf v}|}\\
    \dot{\boldsymbol\lambda}_{\mathbf{r}} = \mu \left(\frac{\boldsymbol{\lambda}_{\mathbf v}}{r^3} - 3(\boldsymbol \lambda_{\mathbf v}\cdot\mathbf r)\frac{\mathbf r}{r^5} \right) - \boldsymbol\omega\times\boldsymbol\omega\times\boldsymbol \lambda_{\mathbf{v}}\ \\
    \dot{\boldsymbol\lambda}_{\mathbf{v}} = - \boldsymbol{\lambda}_{\mathbf r} +2\boldsymbol\omega\times\boldsymbol\lambda_{\mathbf{v}}
\end{array}
\right.
\end{equation}
Since we consider this to be a free time problem, a trajectory also need to fulfill the $\mathcal{H}|_{t=t_f}=0$ condition in order to be optimal.
Let's find one solution, which we'll refer to as the "nominal trajectory" for this problem in the rest of the paper. We introduce a shooting function to solve the Two Points Boundary Value Problem (TPBVP):
\begin{equation}
\label{eq:shooting_transfer}
\phi(\boldsymbol \lambda_{\mathbf{r}_0}, \boldsymbol \lambda_{\mathbf{v}_0}, \lambda_J, t_f) = \left\{\mathbf r_f - \mathbf r_t,  \mathbf v_f - \mathbf v_t, \mathcal H_f,||\boldsymbol \lambda || -1 \right\}
\end{equation}
where $\boldsymbol \lambda_{\mathbf{r}_0}, \boldsymbol \lambda_{\mathbf{v}_0}$ are the initial co-states values and $t_f$ is the time-of-flight. The final conditions $\mathbf r_f$, $\mathbf v_f$, and $\mathcal H_f$ are computed by propagating Eq.\ref{eq:augdyn} from the initial conditions until $t_f$. 
We find a root of $\phi$ using the sequential quadratic programming solver SNOPT \cite{gill2005snopt}. The constraint on the magnitude of the initial co-states $||\boldsymbol \lambda || -1$ is not strictly necessary, we use it here as it improves numerical stability. As described in \cite{dario_seb_gcnet}, the existence of multiple roots for Eq.\ref{eq:shooting_transfer} corresponds to the presence of local minima. While not rigorous, we circumvent this problem by solving this problem using different initial guesses for the numerical solver, thereby increasing our confidence that our solution corresponds to the optimal strategy.
The nominal trajectory for this optimal control problem has a time-of-flight of $t_f^*=4.62$ years, see Fig.~\ref{fig:nominal_transfer}. All values related to this problem are listed in App.\ref{app:1}.
\subsubsection*{Mass-optimal landing on asteroid}
\begin{figure*}[tb]
   \centering
\includegraphics[width=0.7\textwidth]{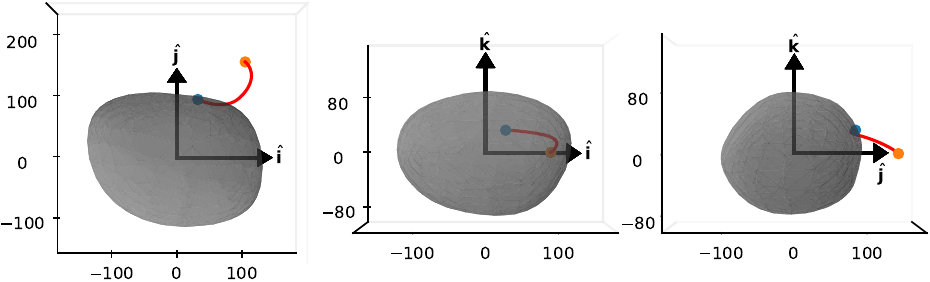}
\caption{Landing on Psyche shown in rotating frame $\mathcal R$. Axis unit is km.}
\label{fig:nominal_landing}
\end{figure*}

We also consider a mass-optimal landing on the asteroid Psyche.
Let $\mathcal R = \left[\hat{\mathbf i}, \hat{\mathbf j}, \hat{\mathbf k}\right]$ be a rotating frame with angular velocity $\boldsymbol \omega \hat {\mathbf k}$ such that the asteroid remains stationary in $\mathcal R$. The dynamics is described by the following ordinary differential equations:
\begin{equation}
\label{eq:dyn_landing}
\left\{ 
\begin{array}{l}
    \dot{x} = v_x \\
    \dot{y} = v_y \\
    \dot{z} = v_z \\
    \dot{v}_x = -\frac{\mu}{r^3}x + 2 \omega v_y + \omega^2 x +u\frac{c_1}{m} i_x \\
    \dot{v}_y = -\frac{\mu}{r^3}y - 2 \omega v_x + \omega^2 y +u\frac{c_1}{m} i_y \\
    \dot{v}_z = -\frac{\mu}{r^3}z +u\frac{c_1}{m} i_z \\
    \dot{m} = -u\frac{c_1}{I_{sp}g_0}
\end{array}
\right.
\end{equation}
The state vector $\mathbf{x}_L$ (subscript $\Box_L$ to indicate the "landing" optimal control problem) contains the position $\mathbf{r}=[x,y,z]$, velocity $\mathbf{v}=[v_x,v_y,v_z]$ and mass $m$. The position $\mathbf{r}$ and velocity $\mathbf{v}$ are both defined in the rotating frame $\mathcal R$. Note that $r = \sqrt{x^2+y^2+z^2}$.
The system is controlled by the thrust direction described by the unit vector $\mathbf{\hat{t}} = [i_x, i_y, i_z]$ and the throttle $u \in [0,1]$.
This is a free-time optimal control problem for which we need to find the controls $u(t)$ and $\mathbf{\hat{t}}(t)$, where $t\in [t_0,t_f]$, such that, under the dynamics described by Eq.\ref{eq:dyn_landing}, the state is steered from any initial state $\mathbf{r}_0$, $\mathbf{v}_0$, $m_0$ to the desired target state $\mathbf{r}_t$, $\mathbf{v}_t$ and final mass $m_f$ (which is left free). To avoid having to immediately solve the mass-optimal control problem we follow the steps laid out in \cite{izzo2021real,bertrand2002new} and introduce the following cost function to minimize:
\begin{equation}
    \label{eq:cost_landing}
    J(u(t), t_f) = \int_{0}^{t_f}\left\{u - \epsilon \log{[u(1-u)]} \right\} \mathrm{d}t
\end{equation}
where the continuation parameter $\epsilon$ and the logarithmic barrier allow us to smooth out the problem and keep $u\in[0,1]$ in the desired bounds. The mass-optimal problem corresponds to $\lim_{{\epsilon \to 0}} \quad J(u(t), t_f) = (m_0-m_f)\cdot\frac{c_1}{I_{sp}g_0}$ (substitute Eq.\ref{eq:dyn_landing} in Eq.\ref{eq:cost_landing}) and is very difficult to solve without a good initial guess. We bypass this issue by first solving the problem with $\epsilon=1$ and use this solution as an initial guess for a slightly smaller $\epsilon$, repeating this cycle until we reach $\epsilon<10^{-6}$.
Let's find the necessary conditions for optimality using Pontryagin's Maximum Principle \cite{pontryagin}. We define the Hamiltonian:
\begin{equation}
\label{eq:hamiltonian2}
\begin{split}
    \mathcal H(\mathbf{r},\mathbf{v},m,\boldsymbol{\lambda}_{\mathbf r},\boldsymbol{\lambda}_{\mathbf v},\lambda_m,u,\mathbf{\hat{t}})= \boldsymbol{\lambda}_{\mathbf r}\cdot \mathbf{v}+ \\ \boldsymbol{\lambda}_{\mathbf v}\cdot\bigg(-\frac{\mu}{r^3}\mathbf{r}-2\boldsymbol \omega\times\mathbf v-\boldsymbol \omega\times\boldsymbol \omega\times\mathbf r+u\frac{c_1}{m}\mathbf{\hat{t}}\bigg) \\ + \lambda_m \bigg(-u\frac{c_1}{I_{sp}g_0}\bigg) + u - \epsilon \cdot \log{[u(1-u)]}
\end{split}
\end{equation}
where $\boldsymbol{\lambda}_{\mathbf r}$, $\boldsymbol{\lambda}_{\mathbf v}$ and $\lambda_m$ are the co-sates functions. Note that we drop the dependence on time for brevity. The optimal thrust direction $\mathbf{\hat{t}}^*$ and $u^*$ both need to minimize the Hamiltonian, hence:
\begin{alignat}{2}
    & \mathbf{\hat{t}}^*=-\frac{ \boldsymbol{\lambda}_{\mathbf v}}{|\boldsymbol{\lambda}_{\mathbf v}|} \quad , \quad  && u^* = \frac{2\epsilon}{2\epsilon + SF + \sqrt{4\epsilon^2 + SF^2}}
\end{alignat}
where $SF$ is a switching function whose zero-crossings correspond to switches between minimal ($u=0$) and maximal throttle ($u=1$):
\begin{equation}
\label{eq:switching}
SF = \boldsymbol{\lambda}_{\mathbf v} \frac{c_1}{m}\mathbf{\hat{t}}^* -\lambda_m\cdot \frac{c_1}{I_{sp}g_0} + 1
\end{equation}
The augmented system of equations is again obtained by taking the derivatives of the Hamiltonian with respect to $\dot {\mathbf x}_L = \frac{\partial \mathcal H}{\partial \boldsymbol \lambda}$ and $\dot {\boldsymbol \lambda} = - \frac{\partial \mathcal H}{\partial \mathbf x_L}$:
\begin{equation}
\label{eq:augdyn2}
\left\{ 
\begin{array}{l}
    \dot{\mathbf{r}} = \mathbf{v} \\
    \dot{\mathbf{v}} =  -\frac{\mu}{r^3}\mathbf{r}-2\boldsymbol \omega\times\mathbf v-\boldsymbol \omega\times\boldsymbol \omega\times\mathbf r-u^*\frac{c_1}{m}\frac{ \boldsymbol{\lambda}_{\mathbf v}}{|\boldsymbol{\lambda}_{\mathbf v}|}\\
    \dot{m} = -u^*\frac{c_1}{I_{sp}g_0} \\ 
    \dot{\boldsymbol\lambda}_{\mathbf{r}} = \mu \left(\frac{\boldsymbol{\lambda}_{\mathbf v}}{r^3} - 3(\boldsymbol \lambda_{\mathbf v}\cdot\mathbf r)\frac{\mathbf r}{r^5} \right) - \boldsymbol\omega\times\boldsymbol\omega\times\boldsymbol \lambda_{\mathbf v}\ \\
    \dot{\boldsymbol\lambda}_{\mathbf{v}} = - \boldsymbol{\lambda}_{\mathbf r} +2\boldsymbol\omega\times\boldsymbol\lambda_{\mathbf v} \\
    \dot{\lambda}_{m} = -\frac{c_1 u^*}{m^2} \boldsymbol{\lambda}_{\mathbf v} \cdot \frac{ \boldsymbol{\lambda}_{\mathbf v}}{|\boldsymbol{\lambda}_{\mathbf v}|}
\end{array}
\right.
\end{equation}
Since this is free time problem, we need to add the condition $\mathcal{H}|_{t=t_f}=0$ and to leave the final mass $m_f$ free we need the transversality condition $\lambda_{m}|_{t=t_f}=0$. We introduce a shooting function to solve the TPBVP:
\begin{equation}
\label{eq:shooting_landing}
\phi(\boldsymbol \lambda_{\mathbf{r}_0}, \boldsymbol \lambda_{\mathbf{v}_0}, \lambda_{m_0}, t_f) = \left\{\mathbf r_f - \mathbf r_t,  \mathbf v_f - \mathbf v_t, \mathcal H_f,\lambda_{m_f} \right\}
\end{equation}
where $\boldsymbol \lambda_{\mathbf{r}_0}, \boldsymbol \lambda_{\mathbf{v}_0}, \lambda_{m_0}$ are the initial co-states values and $t_f$ is the time-of-flight. The final conditions $\mathbf r_f$, $\mathbf v_f$, $\mathcal H_f$ and $\lambda_{m_f}$ are computed by propagating Eq.\ref{eq:augdyn} from the initial conditions until $t_f$. As explained for the time-optimal transfer, we solve this TPBVP with multiple restarts (different initial guesses for the root solver) such as to increase our confidence that our solution is the optimal landing strategy.
The nominal trajectory for this optimal control problem has a time-of-flight of $t_f^*=38$ min,  see Fig.\ref{fig:nominal_landing}. All values related to this problem are listed in App.\ref{app:1}.
\subsection{Behavioural cloning}\label{subsec:BC}
\begin{figure}[tb]
\centering
\def\layersep{2cm}
\def\nodesep{0.3cm}

\begin{tikzpicture}[shorten >=1pt,->,draw=black!50, node distance=\layersep, scale=0.7]
    \tikzstyle{every pin edge}=[<-,shorten <=1pt]
    \tikzstyle{neuron}=[circle,fill=black!25,minimum size=5pt,inner sep=0pt]
    \tikzstyle{input neuron}=[neuron, fill=gray!50];
    \tikzstyle{output neuron}=[neuron, fill=red!50];
    \tikzstyle{output neuron2}=[neuron, fill=green!50];

    \tikzstyle{hidden neuron softplus}=[neuron, fill=blue!50];

    \tikzstyle{annot} = [text width=4em, text centered]

    \node[input neuron, pin=left:$x$] (I-1) at (0,-\nodesep*1*1.63) {};
    \node[input neuron, pin=left:$y$] (I-2) at (0,-\nodesep*2*1.63) {};
    \node[input neuron, pin=left:$z$] (I-3) at (0,-\nodesep*3*1.63) {};
    \node[input neuron, pin=left:$v_x$] (I-4) at (0,-\nodesep*4*1.63) {};
    \node[input neuron, pin=left:$v_y$] (I-5) at (0,-\nodesep*5*1.63) {};
    \node[input neuron, pin=left:$v_z$] (I-6) at (0,-\nodesep*6*1.63) {};


    \foreach \name / \y in {1,...,14}
        \path[yshift=0.5cm]
            node[hidden neuron softplus] (H1-\name) at (\layersep,-\nodesep*\y) {};

    \foreach \name / \y in {1,...,14}
        \path[yshift=0.5cm]
            node[hidden neuron softplus] (H2-\name) at (2*\layersep,-\nodesep*\y) {};
            
    \foreach \name / \y in {1,...,14}
        \path[yshift=0.5cm]
            node[hidden neuron softplus] (H3-\name) at (2.5*\layersep,-\nodesep*\y) {};


    \node[output neuron2, pin={[pin edge={->}]right:$i_x$}, right of=H3-3] (O-1) at (2.3*\layersep,-1.2)  {};
    \node[output neuron2, pin={[pin edge={->}]right:$i_y$}, right of=H3-3] (O-2) at (2.3*\layersep,-1.8)  {};
    \node[output neuron2, pin={[pin edge={->}]right:$i_z$}, right of=H3-3] (O-3) at (2.3*\layersep,-2.4)  {};


    \foreach \source in {1,...,6}
        \foreach \dest in {1,...,14}
            \path (I-\source) edge (H1-\dest);
            
    \foreach \source in {1,...,14}
        \foreach \dest in {1,...,14}
            \path (H1-\source) edge (H2-\dest);

    \path[thick] ([xshift=.1cm] H2-8.east) edge[-,loosely dotted] ([xshift=.1cm] H3-8.east);

    \foreach \source in {1,...,14}
        \foreach \dest in {1,...,3}
            \path (H3-\source) edge (O-\dest);

    \pgfmathsetmacro{\scale}{0.2}
    \pgfmathsetmacro{\Ox}{10}
    \pgfmathsetmacro{\Oy}{4}

    \draw[scale=\scale, shift = {(\Ox, \Oy)}] (-3,0) -- (3,0) node[right] {};
    \draw[scale=\scale, shift = {(\Ox, \Oy)}] (0,-1) -- (0,3) node[above] {softplus};
    \draw[scale=\scale,domain=-3:3,smooth,variable=\x,blue, shift = {(\Ox, \Oy)} ] plot ({\x},{ln(1+exp(\x)});
    
    \draw[scale=\scale, shift = {(\Ox+10, \Oy)}] (-3,0) -- (3,0) node[right] {};
    \draw[scale=\scale, shift = {(\Ox+10, \Oy)}] (0,-1) -- (0,3) node[above] {};
    \draw[scale=\scale,domain=-3:3,smooth,variable=\x,blue, shift = {(\Ox+10, \Oy)} ] plot ({\x},{ln(1+exp(\x)});
    
    \draw[scale=\scale, shift = {(\Ox+15, \Oy)}] (-1,0) -- (3,0) node[right] {};
    \draw[scale=\scale, shift = {(\Ox+15, \Oy)}] (0,-1) -- (0,3) node[above] {};
    \draw[scale=\scale,domain=-2:3,smooth,variable=\x,blue, shift = {(\Ox+15, \Oy)} ] plot ({\x},{ln(1+exp(\x)});
    
    \draw[scale=\scale, shift = {(\Ox+30, \Oy-5)}] (-1,0) -- (3,0) node[right] {};
    \draw[scale=\scale, shift = {(\Ox+30, \Oy-5)}] (0,-1) -- (0,3) node[above] {linear};
    \draw[scale=\scale,domain=-2:3,smooth,variable=\x,green, shift = {(\Ox+30, \Oy-5)} ] plot ({\x},{(\x)});

\end{tikzpicture}
\begin{tikzpicture}[shorten >=1pt,->,draw=black!50, node distance=\layersep, scale=0.7]
    \tikzstyle{every pin edge}=[<-,shorten <=1pt]
    \tikzstyle{neuron}=[circle,fill=black!25,minimum size=5pt,inner sep=0pt]
    \tikzstyle{input neuron}=[neuron, fill=gray!50];
    \tikzstyle{output neuron}=[neuron, fill=red!50];
    \tikzstyle{output neuron2}=[neuron, fill=green!50];

    \tikzstyle{hidden neuron softplus}=[neuron, fill=blue!50];

    \tikzstyle{annot} = [text width=4em, text centered]

    \node[input neuron, pin=left:$x$] (I-1) at (0,-\nodesep*1*1.63) {};
    \node[input neuron, pin=left:$y$] (I-2) at (0,-\nodesep*2*1.63) {};
    \node[input neuron, pin=left:$z$] (I-3) at (0,-\nodesep*3*1.63) {};
    \node[input neuron, pin=left:$v_x$] (I-4) at (0,-\nodesep*4*1.63) {};
    \node[input neuron, pin=left:$v_y$] (I-5) at (0,-\nodesep*5*1.63) {};
    \node[input neuron, pin=left:$v_z$] (I-6) at (0,-\nodesep*6*1.63) {};
    \node[input neuron, pin=left:$m$] (I-7) at (0,-\nodesep*7*1.63) {};


    \foreach \name / \y in {1,...,15}
        \path[yshift=0.5cm]
            node[hidden neuron softplus] (H1-\name) at (\layersep,-\nodesep*\y) {};

    \foreach \name / \y in {1,...,15}
        \path[yshift=0.5cm]
            node[hidden neuron softplus] (H2-\name) at (2*\layersep,-\nodesep*\y) {};
            
    \foreach \name / \y in {1,...,15}
        \path[yshift=0.5cm]
            node[hidden neuron softplus] (H3-\name) at (2.5*\layersep,-\nodesep*\y) {};

    \node[output neuron, pin={[pin edge={->}]right:$u$}, right of=H3-3] (O-1) at (2.3*\layersep,-0.2)  {};
    \node[output neuron2, pin={[pin edge={->}]right:$i_x$}, right of=H3-3] (O-2) at (2.3*\layersep,-3)  {};
    \node[output neuron2, pin={[pin edge={->}]right:$i_y$}, right of=H3-3] (O-3) at (2.3*\layersep,-3.4)  {};
    \node[output neuron2, pin={[pin edge={->}]right:$i_z$}, right of=H3-3] (O-4) at (2.3*\layersep,-3.8)  {};

    \foreach \source in {1,...,7}
        \foreach \dest in {1,...,15}
            \path (I-\source) edge (H1-\dest);
            
    \foreach \source in {1,...,15}
        \foreach \dest in {1,...,15}
            \path (H1-\source) edge (H2-\dest);

    \path[thick] ([xshift=.1cm] H2-8.east) edge[-,loosely dotted] ([xshift=.1cm] H3-8.east);

    \foreach \source in {1,...,15}
        \foreach \dest in {1,...,4}
            \path (H3-\source) edge (O-\dest);

    \pgfmathsetmacro{\scale}{0.2}
    \pgfmathsetmacro{\Ox}{10}
    \pgfmathsetmacro{\Oy}{4}

    \draw[scale=\scale, shift = {(\Ox, \Oy)}] (-3,0) -- (3,0) node[right] {};
    \draw[scale=\scale, shift = {(\Ox, \Oy)}] (0,-1) -- (0,3) node[above] {softplus};
    \draw[scale=\scale,domain=-3:3,smooth,variable=\x,blue, shift = {(\Ox, \Oy)} ] plot ({\x},{ln(1+exp(\x)});
    
    \draw[scale=\scale, shift = {(\Ox+10, \Oy)}] (-3,0) -- (3,0) node[right] {};
    \draw[scale=\scale, shift = {(\Ox+10, \Oy)}] (0,-1) -- (0,3) node[above] {};
    \draw[scale=\scale,domain=-3:3,smooth,variable=\x,blue, shift = {(\Ox+10, \Oy)} ] plot ({\x},{ln(1+exp(\x)});
    
    \draw[scale=\scale, shift = {(\Ox+15, \Oy)}] (-1,0) -- (3,0) node[right] {};
    \draw[scale=\scale, shift = {(\Ox+15, \Oy)}] (0,-1) -- (0,3) node[above] {};
    \draw[scale=\scale,domain=-2:3,smooth,variable=\x,blue, shift = {(\Ox+15, \Oy)} ] plot ({\x},{ln(1+exp(\x)});
    
    \draw[scale=\scale, shift = {(\Ox+30, \Oy-2)}] (-1,0) -- (3,0) node[right] {};
    \draw[scale=\scale, shift = {(\Ox+30, \Oy-2)}] (0,-1) -- (0,3) node[above] {sigmoid};
    \draw[scale=\scale,domain=-2:3,smooth,variable=\x,red, shift = {(\Ox+30, \Oy-2)} ] plot ({\x},{1/(1+exp(-\x * 2))});
    
    \draw[scale=\scale, shift = {(\Ox+30, \Oy-15)}] (-1,0) -- (3,0) node[right] {};
    \draw[scale=\scale, shift = {(\Ox+30, \Oy-15)}] (0,-1) -- (0,3) node[above] {linear};
    \draw[scale=\scale,domain=-2:3,smooth,variable=\x,green, shift = {(\Ox+30, \Oy-15)} ] plot ({\x},{(\x)});
    
\end{tikzpicture}
\caption{G\&CNET architectures: transfer (top) and landing (bottom). Adapted from \cite{izzo2021real}. \label{fig:ffnn}}

\end{figure}
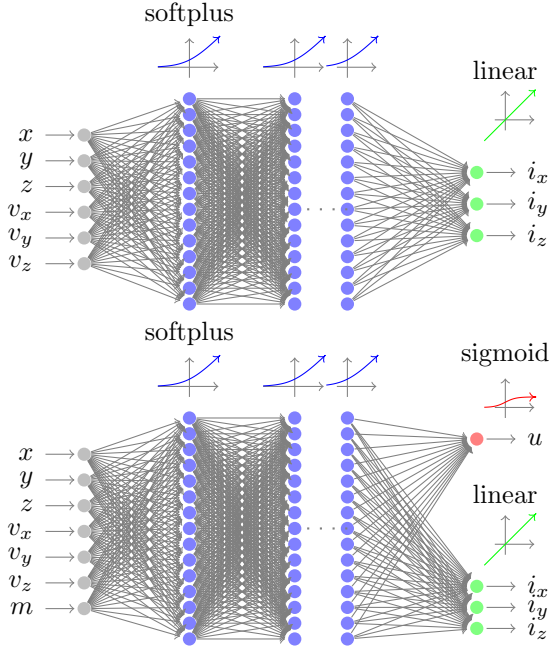

We train two separate neural models to represent to optimal policies for each problem as a function of the spacecraft state $\mathbf{x}$. The resulting neural state feedback is called a G\&CNET: $\mathcal N_{\text{transfer}}(\mathbf x_T) = \mathbf{\hat{t}}^*$ for the transfer and $\mathcal N_{\text{landing}}(\mathbf x_L) =[u^*,\mathbf{\hat{t}}^*]$ for the landing. These simple feedforward neural networks can then be used in Eq.\ref{eq:dyn_transfer} and Eq.\ref{eq:dyn_landing} respectively to simulate the spacecraft dynamics. The network architectures are shown in Fig.\ref{fig:ffnn}, for both problems we use 3 hidden layers, each with 128 neurons. We use softplus activation functions for the hidden layers, allowing us to obtain a continuous and differentiable representation of the optimal controls. To avoid saturation issues during training we use linear output activation functions, except for the throttle in the landing problem where we use a sigmoid activation function to keep $u$ bounded between $[0,1]$.
\subsubsection*{Training datasets}
We generate training datasets for both optimal control problems by leveraging a data augmentation technique called the "Backward Generation of Optimal Examples" (BGOE) \cite{izzo2021real,dario_seb_gcnet}. This technique exploits the fact that any solution to the augmented systems of equations (Eq.\ref{eq:augdyn} for the transfer and Eq.\ref{eq:augdyn2} for the landing problem) which satisfies the necessary conditions for optimality is a local optimal trajectory which can be used to learn from. The basic premise of the BGOE is that once a nominal solution is found, one can perturb the final co-states of the augmented system by some carefully crafted vector $\boldsymbol\Delta$: 
\begin{equation}\label{eq:pert}
   \boldsymbol \lambda_{f}^{+} = \boldsymbol \lambda_{f} + \boldsymbol \lambda_{f}\cdot\boldsymbol\Delta
\end{equation}
where $\boldsymbol\Delta$ needs to be chosen such that the necessary conditions for optimality are still satisfied.
In the case of the transfer, each element in $\boldsymbol\Delta$ is a number uniformly sampled in $\mathcal U(-\delta, \delta)$ except for $\lambda_J$ which we use to satisfy the free time condition $\mathcal{H}_f=0$. For the landing problem we also sample each element in $\mathcal U(-\delta, \delta)$ except for $\lambda_{m_f}=0$ which we leave unchanged (free final mass transversality condition) and we use the final mass $m_f$ to satisfy the free time condition $\mathcal{H}_f=0$ using a root solver and the final mass $m_f$ of the nominal trajectory as initial guess. $\boldsymbol \lambda_{f}^{+}$ can then be used to back-propagate the augmented system of equation in time. 
For small perturbations $\delta$, this will result in a new optimal trajectory with the same final states as the nominal trajectory (except for the final mass $m_f$ which is left free in the landing problem) and different initial conditions. While we cannot directly chose the initial conditions, the BGOE allows us to generate hundred thousand optimal trajectories at a fraction of the computational cost one would incur if one had to solve each TPBVP individually \cite{izzo2021real, dario_seb_gcnet}.

\begin{figure}[tb]
   \centering
\includegraphics[width=\columnwidth]{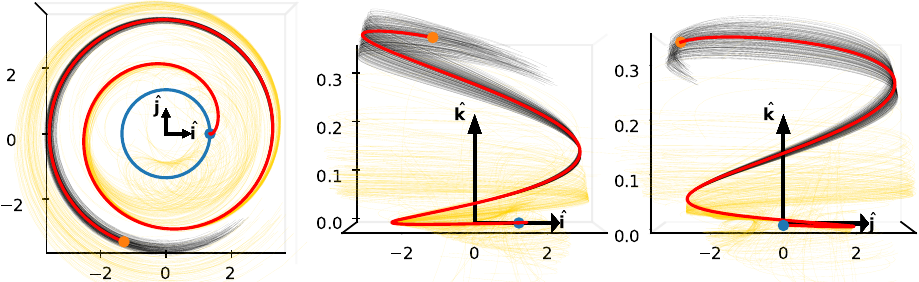}
\caption{Bundle of 400 optimal trajectories. Co-state perturbation $\delta$:  $1\permil$ (black), $8\%$ (gold).  Interplanetary transfer shown in rotating frame $\mathcal F$. Axis unit is AU.}\label{fig:bundle_transfer}
\end{figure}

For the transfer we use 400,000 optimal trajectories, each of which is sampled in 100 points equally distance in time, resulting in 40,000,000 optimal state-action pairs to learn from. We show a portion of this dataset in Fig.\ref{fig:bundle_transfer}. In our experiments, we found that decreasing the correlation with the nominal trajectory is crucial to successfully train the G\&CNET. We do this by randomly sampling the back-propagation time $a(1+c)t^*_{f_{nom}}$ where $t^*_{f_{nom}}$ is the optimal time-of-flight of the nominal trajectory, $a=1$ and $c\in[0,0.07]$. We create two bundles of trajectories, one with a small perturbation size $\delta = 1\permil$, hence it closely follows the nominal trajectory, and one with a much larger perturbation size $\delta = 8\%$. As explained in \cite{dario_seb_gcnet}, while these trajectories are far away from the nominal trajectory, they were crucial for successful training.
We also generate a separate dataset of 1,000 trajectories with $\delta = 1\permil$ which will be used to in Subsec.\ref{sec:results_and_discussion}.

\begin{figure}[tb]
   \centering
\includegraphics[width=\columnwidth]{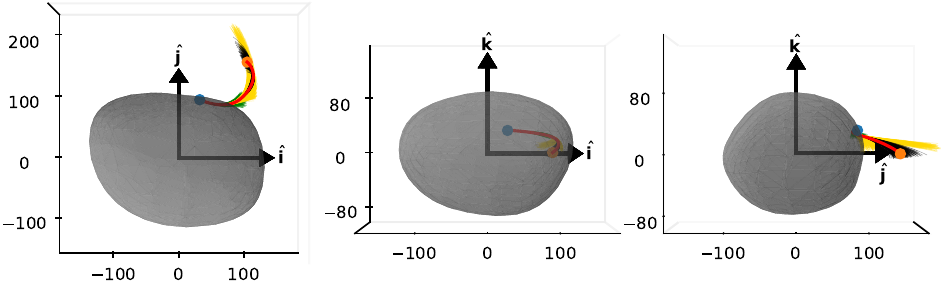}
\caption{Bundle of 2,000 optimal trajectories. Co-state perturbation $\delta$: $1\permil$ (black), $8\permil$ (gold), $2\%$ (green). Landing on Psyche shown in rotating frame $\mathcal R$. Axis unit is km.}
\label{fig:bundle_landing}
\end{figure}

For the landing we use 300,000 optimal trajectories, each of which is sampled in 100 points equally distance in time, resulting in 30,000,000 optimal state-action pairs to learn from, see a portion of the dataset in Fig.\ref{fig:bundle_landing}. We use $c\in[0,0.05]$ and create three bundles of different perturbation size and only back-propagated until a fraction of $t^*_{f_{nom}}$: $\delta = 1\permil$ ($a=1$), $\delta = 8\permil$ ($a=0.8$) and $\delta = 2\%$ ($a=0.5$).
We also generate a separate dataset of 1,000 trajectories with $\delta = 0.5\permil$ ($a=1$) for Subsec.\ref{sec:results_and_discussion}.

Central to our work is the \textit{heyoka} Python library \cite{biscani2021revisiting}, which we use for all numerical propagation. 
This Taylor based method allows us to quickly integrate our equations with a numerical tolerance set to machine level, i.e. $10^{-16}$. In addition, as we will see in Sec.\ref{subsec:NeuralODEs}, the library also allows for a seamless integration of feed forward neural networks in our ODEs and automatic differentiation.
\subsubsection*{Training procedure}
\begin{figure*}[!t]
  \centering
  \includegraphics[width=0.425\textwidth]{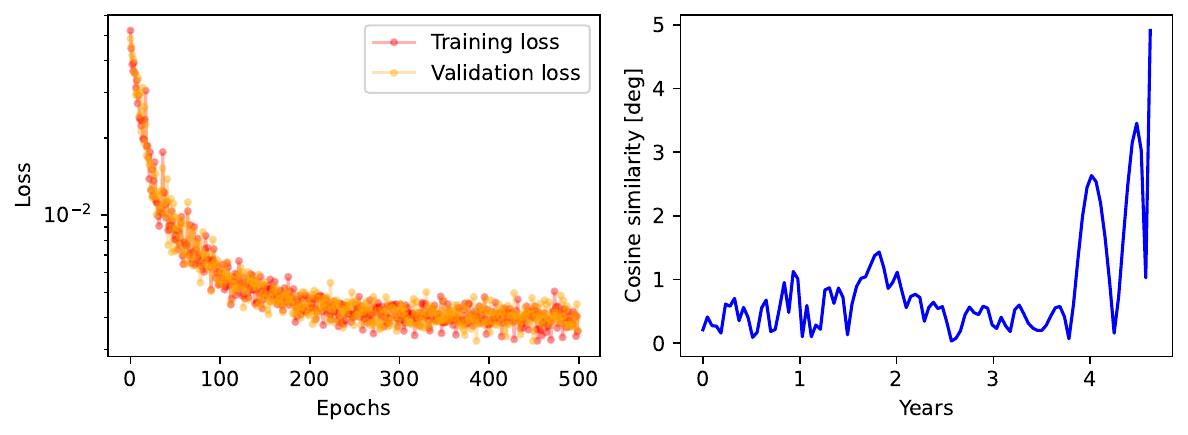}
  \hfill
  \includegraphics[width=0.535\textwidth]{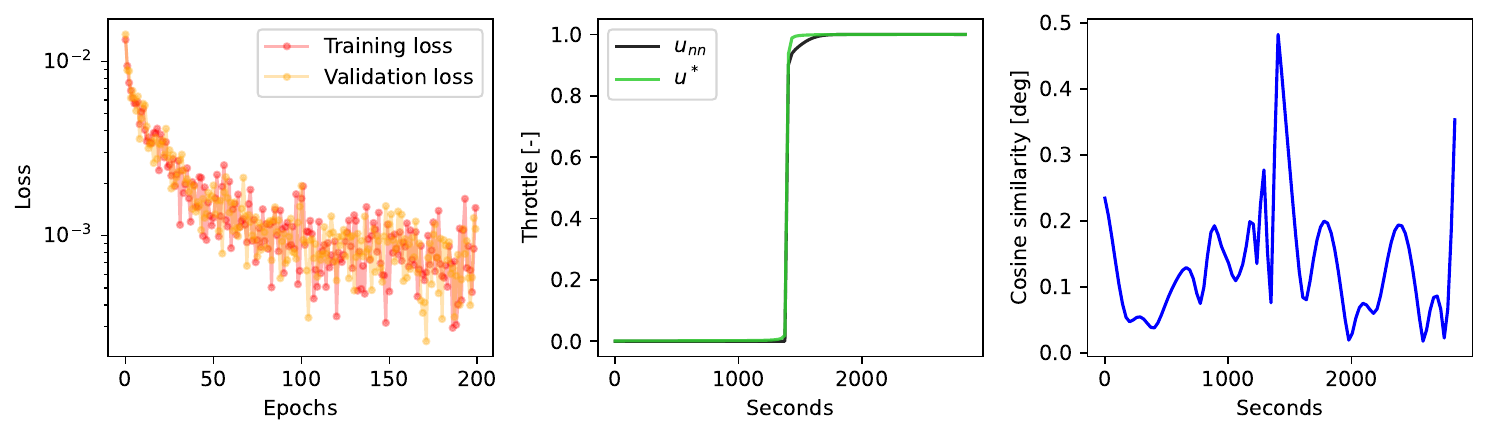}
  \caption{Loss during training and cosine similarity between the estimated thrust direction and the ground truth over one optimal trajectory in the validation dataset. Transfer (left) and landing (right). For the landing we also show the estimated throttle versus ground truth.}
  \label{fig:losses_figs}
\end{figure*}

The generated datasets are split up into $80\%$ training data and $20\%$ validation data. We train the G\&CNET for the transfer problem over $p=500$ epochs using an initial learning rate of $\alpha=5\cdot 10^{-5}$ with the Adam optimizer \cite{kingma2014adam} and no weight decay. We also use a scheduler which decreases the learning rate by a factor of $90\%$ whenever the loss does not improve over $p=10$ epochs when evaluated on the validation dataset.
The loss function is computed using the cosine similarity of the estimated thrust direction $\mathbf{\hat{t}}_{nn}$ and the ground truth $\mathbf{\hat{t}}^*$, thereby allowing the network to solely focus on the direction and ignoring the norm:
\begin{equation}
\text{cosine\_similarity}\left(\mathbf{\hat{t}}_{nn},\mathbf{\hat{t}}^*\right) = \frac{\mathbf{\hat{t}}^*\cdot \mathbf{\hat{t}}_{nn}}{\mathbf{\vert\hat{t}}^*\vert\vert \mathbf{\hat{t}}_{nn}\vert}
\end{equation}
\begin{equation}\label{eq:loss_transfer}
    \mathcal L_{transfer}\left(\mathbf{\hat{t}}_{nn},\mathbf{\hat{t}}^*\right) = 1-\text{cosine\_similarity}\left(\mathbf{\hat{t}}_{nn}, \mathbf{\hat{t}}^*\right)
\end{equation}
We report the loss over the epochs during training and the error in thrust direction (represented by the cosine similarity) over one trajectory in the validation dataset in Fig.\ref{fig:losses_figs}. Notice how the final part of the transfer (last $0.5$ year) is usually where the largest errors occur, likely due to a lack of training data in the region of space corresponding to the final part of the transfer.

For the landing problem we use exactly the same training setup except that the total amount of epochs is now $p=400$ and the loss function contains an additional term to penalize the Mean Squared Error (MSE) between the estimated throttle $u_{nn}$ and the ground truth $u^*$:
\begin{equation}\label{eq:loss_landing}
\begin{split}
        \mathcal L_{landing}\left(u_{nn},u^*,\mathbf{\hat{t}}_{nn},\mathbf{\hat{t}}^*\right) = \text{MSE}(u_{nn},u^*) + 1\\-\text{cosine\_similarity}\left(\mathbf{\hat{t}}_{nn}, \mathbf{\hat{t}}^*\right)
\end{split}
\end{equation}
Fig.\ref{fig:losses_figs} shows the loss during training, the estimated throttle versus the ground truth and the cosine similarity between the estimated thrust direction and the ground truth over one trajectory in the validation dataset.


\subsection{Neural ODEs}\label{subsec:NeuralODEs}
\begin{algorithm*}[!t]
\caption{Neural ODE fix algorithm.}\label{algo:neural_ode}
\begin{algorithmic}[1]
\State Generate training dataset $D_{BC}$\Comment{Optimal trajectories $[\mathbf{x}^*,\mathbf{u}^*, t^*]$}
\State Train G\&CNET $\mathcal{N}_{\boldsymbol\theta}$ on $D_{BC}$\Comment{Behavioural Cloning}
\State Generate training dataset $D_{T}$\Comment{Optimal trajectories $[\mathbf{x}^*,\mathbf{u}^*, t^*]$}
\State Generate validation dataset $D_V$\Comment{Optimal trajectories $[\mathbf{x}^*,\mathbf{u}^*, t^*]$}
\For{$i=1$ to $N$}
    \State Propagate dynamics $\dot{\mathbf{x}}$ and ODE sensitivities $\frac{d}{dt}\big(    \frac{\partial \mathbf{x}}{\partial \boldsymbol\theta}   \big)$ 
    from $\mathbf{x}^*_{0,i}$ until $t^*_i$ in $D_T$
    \State Compute gradients: $\frac{\partial \mathcal{L}_{N}}{\partial \boldsymbol\theta} = \frac{\partial \mathbf{x}}{\partial \boldsymbol\theta} \frac{\partial \mathcal{L}_{N}}{\partial \mathbf{x}}$
    \State Find optimal learning rate $\alpha$\Comment{Line search or Adam Optimizer}
    \State Perform gradient descent step: $\boldsymbol\theta_{i+1} = \boldsymbol\theta_{i} - \alpha \cdot \frac{\partial \mathcal{L}_N}{\partial \boldsymbol\theta}$
\EndFor
\State \textbf{Return} Best $\mathcal{N}_{\boldsymbol\theta}$ on validation dataset $D_V$
\end{algorithmic}
\end{algorithm*}
We use the term Neural Ordinary Differential Equations (Neural ODEs), as popularized in \cite{neural_ode}, to denote Ordinary Differential Equations which contain an artificial neural network on their right hand side.
To illustrate how these can be used to improve the performance of G\&CNETs let us a consider the generic system $\dot{\mathbf{x}} = f(\mathbf{x}, \mathcal{N}_{\boldsymbol\theta}(\mathbf{x}))$, whose solution $\mathbf{x}(t;\mathbf{x_0},\boldsymbol\theta)$ depends explicitly on the initial conditions $\mathbf{x_0}$ and the network parameters $\boldsymbol\theta$.
Contrary to the behavioural cloning approach, which aims to minimize the approximation error of the optimal control (see loss functions in Eq.\ref{eq:loss_transfer} and Eq.\ref{eq:loss_landing}), imagine that we could update our neural network parameters $\boldsymbol\theta$ such as to decrease some loss $\mathcal{L}(\mathbf{x}(t;\mathbf{x_0},\boldsymbol\theta))$, which instead depends on the current solution of the system.
Let us consider a new loss function $\mathcal{L}_{N}$ which aims to directly minimize the error between the target state $\mathbf{x}_{target}$ and the current solution of our system when evaluated from some initial conditions $\mathbf{x_0}$ until the corresponding optimal time-of-flight $t=t^*$:
\begin{equation}\label{eq:loss_neuralODE}
    \mathcal{L}_{N} = (\mathbf{x}_{target} - \mathbf{x}(t^{*}))^2
\end{equation}
The gradients of the loss with respect to the network parameters can be rewritten as:
\begin{equation}\label{eq:var_equ}
    \frac{\partial \mathcal{L}_{N}}{\partial \boldsymbol\theta} = \frac{\partial \mathbf{x}}{\partial \boldsymbol\theta} \frac{\partial \mathcal{L}_{N}}{\partial \mathbf{x}}
\end{equation}
One can easily find $\frac{\partial \mathcal{L}}{\partial \mathbf{x}}$ analytically, hence we only need to compute  the ODE sensitivities $\frac{\partial \mathbf{x}}{\partial \boldsymbol\theta} $ via the variational equations (or the Pontryagin's adjoint method):
\begin{equation}
    \frac{d}{dt} \Big( \frac{\partial \mathbf{x}}{\partial \boldsymbol\theta}  \Big)= \nabla_{\mathbf{x}} \dot{\mathbf{x}}\cdot \frac{\partial \mathbf{x}}{\partial \boldsymbol\theta} + \frac{\partial \dot{\mathbf{x}}}{\partial \boldsymbol\theta}
\end{equation}
The \textit{heyoka} library \cite{biscani2021revisiting} allows to compute these derivatives very easily, see tutorial\footnote{Tutorial heyoka: \url{ https://bluescarni.github.io/heyoka.py/notebooks/NeuralODEs.html}}. Finally, a gradient descent step can be used to update the neural network weights:
\begin{equation}\label{eq:gradient_descent}
    \boldsymbol\theta_{i+1} = \boldsymbol\theta_{i} - \alpha \cdot \frac{\partial \mathcal{L}_N}{\partial \boldsymbol\theta}
\end{equation}
The training pipeline is provided as pseudo-code in Algorithm~\ref{algo:neural_ode}, let's walk through each step. First we follow the same steps as in the behavioural cloning pipeline: we generate a training dataset $D_{BC}$ (Line 1), and we train the G\&CNET $\mathcal{N}_{\boldsymbol\theta}$ on $D_{BC}$ (Line 2).
We also generate two separate datasets, which will be use to train ($D_T$) and validate ($D_V$) the Neural ODE fix algorithm (Lines 3 and 4). Until now, we always only used the optimal state-action pairs $[\mathbf{x}^*,\mathbf{u}^*]$ from our generated datasets. However solving optimal trajectories also gives the optimal time-of-flight $t^*$, recall that $t_f$ is part of the decision vector in both shooting functions Eq.\ref{eq:shooting_transfer} and Eq.\ref{eq:shooting_landing}. This is crucial here as it allows us to evaluate the following system:
\begin{equation}
\label{eq:dyn_and_odesen}
\left\{ 
\begin{array}{l}
    \dot{\mathbf{x}} = f(\mathbf{x}, \mathcal{N}_{\boldsymbol\theta}(\mathbf{x})) \\
    \frac{d}{dt} \Big( \frac{\partial \mathbf{x}}{\partial \boldsymbol\theta}  \Big)= \nabla_{\mathbf{x}} \dot{\mathbf{x}}\cdot \frac{\partial \mathbf{x}}{\partial \boldsymbol\theta} + \frac{\partial \dot{\mathbf{x}}}{\partial \boldsymbol\theta}
\end{array}
\right.
\end{equation}
from $\mathbf{x}_{0,i}^*$ until $t=t^*_i$ in $D_T$ (Line 6). Since $\mathcal{N}_{\boldsymbol\theta}$ approximates the optimal control policy, errors will accumulate over the trajectory, resulting in a non-zero error to the target state at $t=t^*$ which can be captured by the loss $\mathcal{L}_N$ in Eq.\ref{eq:loss_neuralODE}. The ODE sensitivities to the network parameters $\boldsymbol\theta$ can now be used to compute the gradients of $\mathcal{L}_N$ with respect to these parameters (Line 7). In our experiments we found that using a line search (for instance using \textit{scipy.optimize.minimize\_scalar}) or the Adam Optimizer \cite{kingma2014adam} to find the optimal learning rate $\alpha$ for the gradient descent step helps considerably to stabilize training (Lines 8 and 9). Finally, we return the policy which performs best on the validation dataset $D_V$ (Line 10). Note that in our experiments we distinguish between looping over a single trajectory or batches of trajectories in $D_T$ (Line 5).
\subsection{DAGGER}
\begin{algorithm*}[!t]
\caption{DAGGER algorithm, adapted from \cite{ross2011reduction}.}\label{algo:dagger}
\begin{algorithmic}[1]
\State Generate training dataset $D_{BC}$\Comment{Optimal trajectories $[\mathbf{x}^*,\mathbf{u}^*, t^*]$}
\State Train G\&CNET $\mathcal{N}_{\boldsymbol\theta}$ on $D_{BC}$\Comment{Behavioural Cloning}
\State Generate training dataset $D_{T}$\Comment{Optimal trajectories $[\mathbf{x}^*,\mathbf{u}^*, t^*]$}
\State Generate validation dataset $D_V$\Comment{Optimal trajectories $[\mathbf{x}^*,\mathbf{u}^*, t^*]$}
\For{$i=1$ to $N$}
    \State Propagate $\dot{\mathbf{x}} = f(\mathbf{x}, \mathcal{N}_{\boldsymbol\theta}(\mathbf{x}))$ from $\mathbf{x}^*_{0,i}$ until $t^*_i$ in $D_T$\Comment{Eq.\ref{eq:dyn_transfer} or Eq.\ref{eq:dyn_landing}}
    \State Sample trajectory in $T$ steps
    \For{$j=1$ to $T$}
        \State Solve TPBVP from $\mathbf{x}_j$ with Pontryagin $D_j = \{(\mathbf{x}^*, \mathbf{u}^*)\}$\Comment{Eq.\ref{eq:shooting_transfer} or Eq.\ref{eq:shooting_landing}}
        \State Compute loss $\mathcal{L}_j=\mathcal{L}(\mathcal{N}_{\boldsymbol\theta}(\mathbf{x}_j), \mathbf{u}^*)$\Comment{Eq.\ref{eq:loss_transfer} or Eq.\ref{eq:loss_landing}}
        \If{$\mathcal{L}_j > \text{threshold}$}
            \State Aggregate datasets: $D_{DG} \leftarrow D_{DG} \cup D_j$
        \EndIf
    \EndFor
    \State Train G\&CNET $\mathcal{N}_{\boldsymbol\theta}$ on $D_{DG}$ and $D_{BC}$\Comment{Behavioural Cloning}
\EndFor
\State \textbf{Return} Best $\mathcal{N}_{\boldsymbol\theta}$ on validation dataset $D_V$
\end{algorithmic}
\end{algorithm*}
We compare our approach to the popular 
DAGGER (Dataset Aggregation) algorithm \cite{ross2011reduction}. The idea behind this technique is to let the neural network explore the environment and query an expert (in this case solve the corresponding TPBVP) to obtain the optimal policy, thereby gradually collecting optimal state-action pairs from the states that the network is likely to visit. This technique addresses the issue that the distribution of the initial training data used in behavioural cloning rarely covers the state-space encountered by the network perfectly. Since no new methods or equations need to be introduced, let's run through a concrete example by following the steps laid out in the pseudo-code Algorithm~\ref{algo:dagger}.
Just like for the Neural ODE fix algorithm we first follow the behavioural cloning pipeline by generating a training dataset $D_{BC}$ (Line 1) and training the G\&CNET $\mathcal{N}_{\boldsymbol\theta}$ on $D_{BC}$ (Line 2).
Here we also generate two separate datasets, which will be used to train ($D_T$) and validate ($D_V$) the DAGGER algorithm (Lines 3 and 4). We then propagate the dynamics $\dot{\mathbf{x}} = f(\mathbf{x}, \mathcal{N}_{\boldsymbol\theta}(\mathbf{x}))$ from initial conditions $\mathbf{x}^*_{0,i}$ until the corresponding optimal time-of-flight $t^*_i$ (Line 6). Note that compared to Algorithm~\ref{algo:neural_ode}, here it is not as important to know the optimal time-of-flight. The resulting trajectory is then sampled into $T$ steps (Line 7) and we solve all the TPBVPs starting from the sampled states $\mathbf{x}_j$ until the target state (Line 9). Now we can evaluate how well the network approximates the optimal control policy $\mathbf{u}^*$ at state $\mathbf{x}_j$ and add the corresponding optimal trajectories to a new training dataset $D_{DG}$ when the approximation error surpasses some user-defined threshold (Lines 10, 11 and 12).
Finally the network is trained using Behavioural Cloning on both the old dataset $D_{BC}$ and the new dataset $D_{DG}$ which contains new state-action pairs that are likely to be encountered when deployed and which the network struggles to approximate accurately. In order to prevent catastrophic forgetting, we found that it is necessary to consider a slightly modified loss function when training $\mathcal{N}_{\boldsymbol\theta}$ in Line 13:
\begin{equation}
    \mathcal{L} = \mathcal{L}_{D_{BC}} + 0.1 \cdot \mathcal{L}_{D_{DG}}
\end{equation}
where $\mathcal{L}_{D_{BC}}$ and $\mathcal{L}_{D_{DG}}$ are Eq.\ref{eq:loss_transfer} (transfer) or Eq.\ref{eq:loss_landing} (landing) when evaluated on the optimal state-action pairs of $D_{BC}$ and $D_{DG}$. This new loss function allows us to weigh the contribution of each dataset differently which resulted in more accurate networks when evaluating these on the validation dataset $D_V$ (Line 14).
\section{Results \& Discussion}\label{sec:results_and_discussion}
\begin{figure*}[!t]
  \centering
  \includegraphics[width=0.87\textwidth]{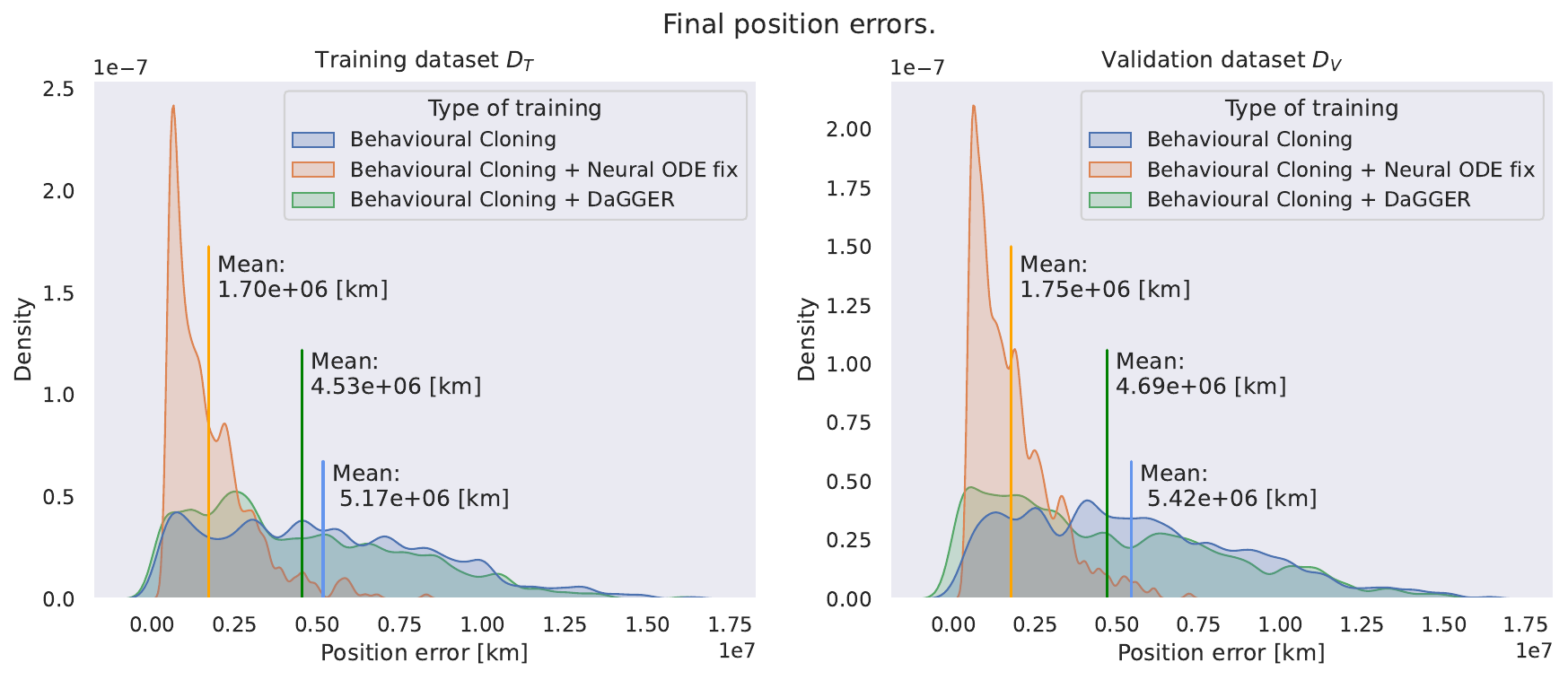}
  \hfill
  \includegraphics[width=0.87\textwidth]{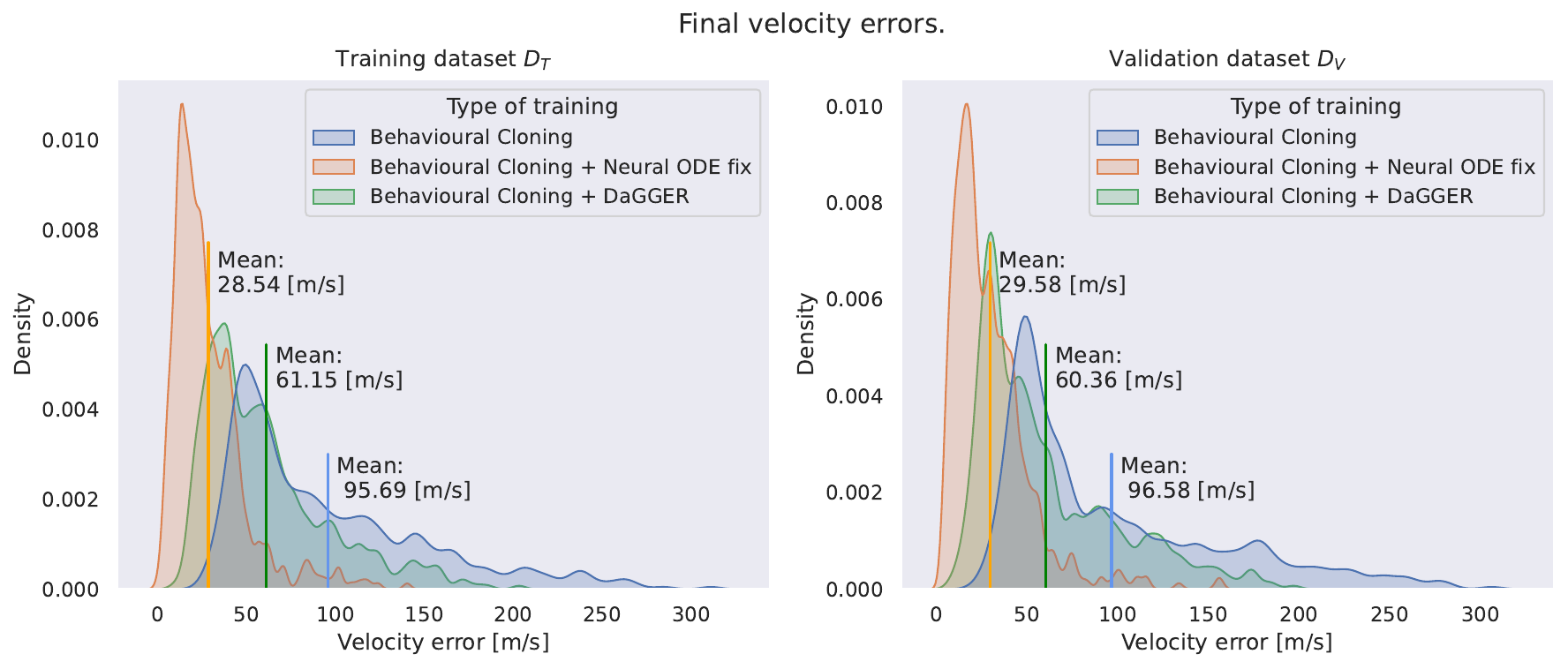}
  \caption{Position and velocity errors over training and testing dataset of initial G\&CNET and refined G\&CNETs for interplanetary transfer problem.}
  \label{fig:closing_the_gap1}
\end{figure*}

\begin{figure*}[!t]
  \centering
  \includegraphics[width=0.87\textwidth]{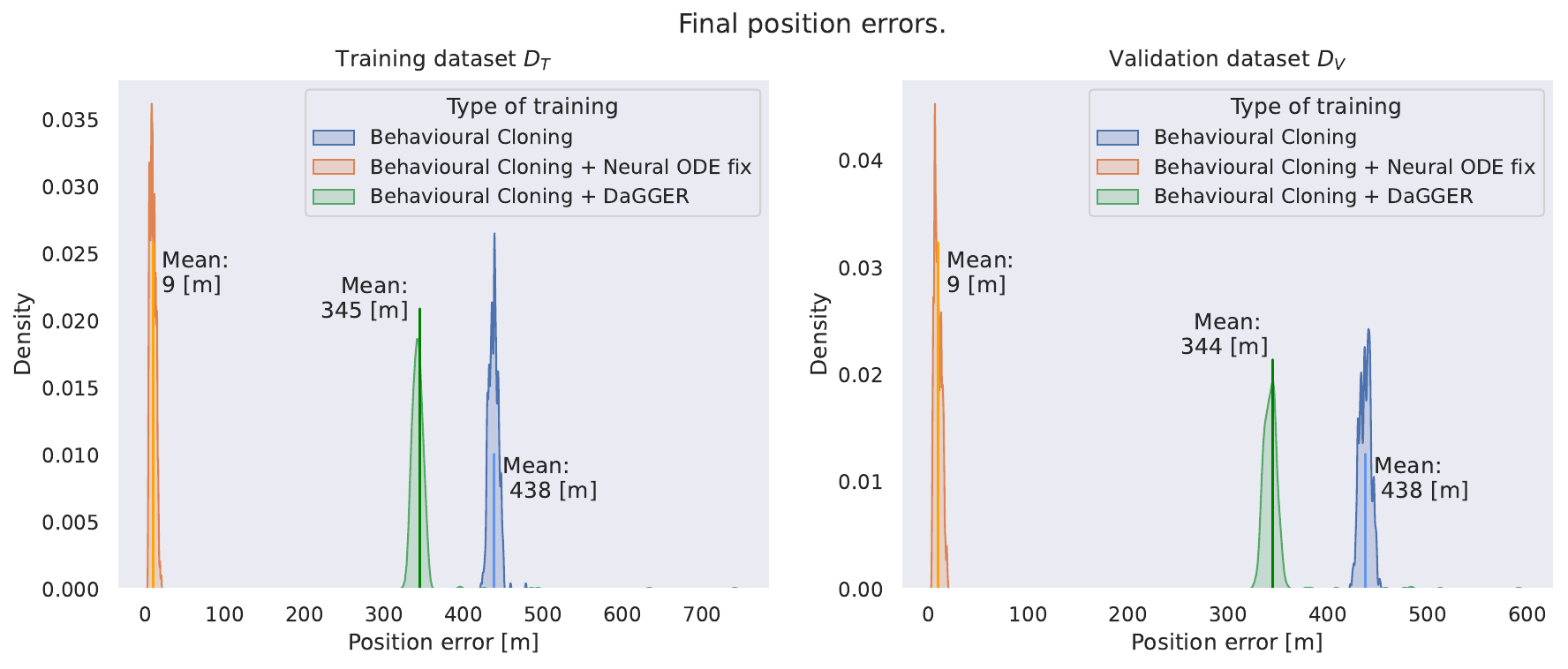}
  \hfill
  \includegraphics[width=0.87\textwidth]{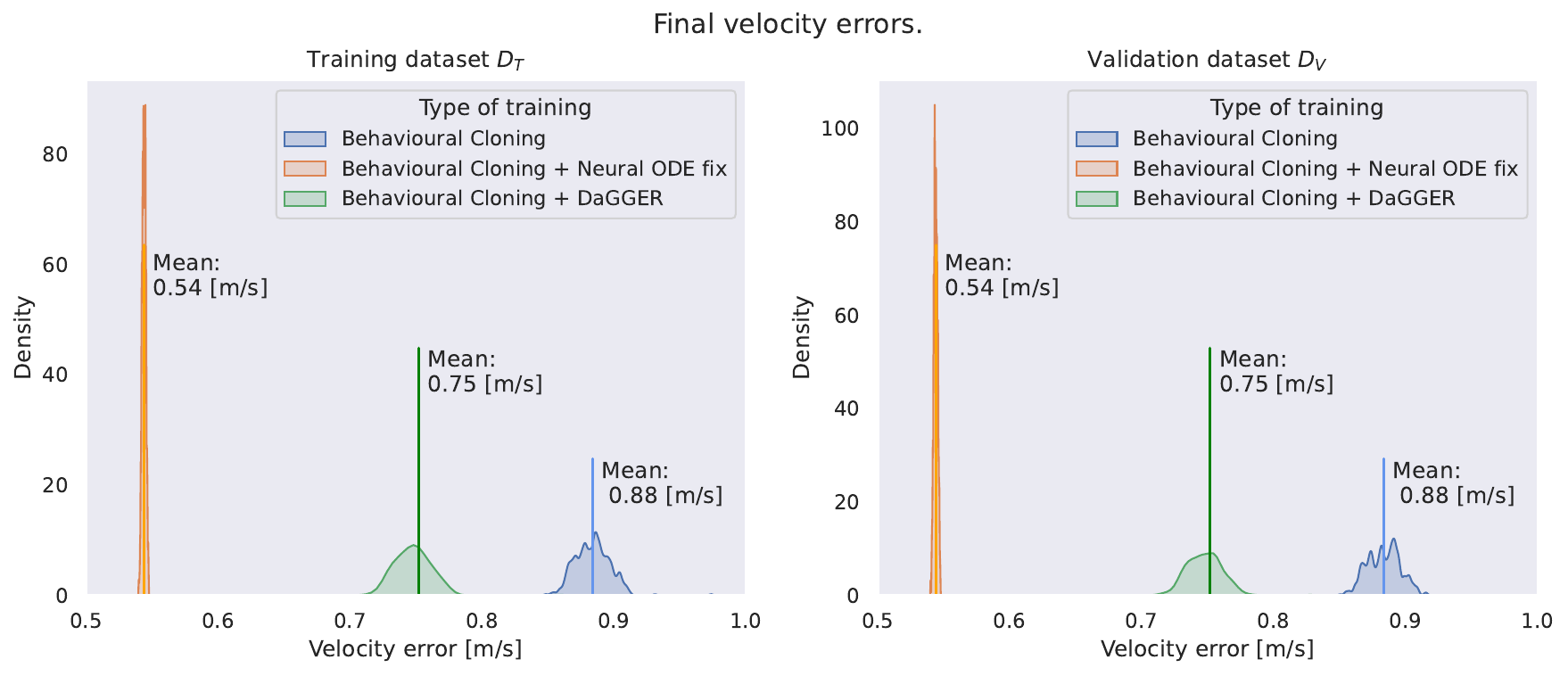}
  \caption{Position and velocity errors over training and testing dataset of initial G\&CNET and refined G\&CNETs for asteroid landing problem.}
  \label{fig:closing_the_gap2}
\end{figure*}

We show the final position and velocity errors on the training dataset $D_T$ and validation dataset $D_V$ (500 trajectories each) for both optimal control problems in Fig.\ref{fig:closing_the_gap1} and Fig.\ref{fig:closing_the_gap2}. In all cases, the G\&CNETs trained solely with behavioural cloning can be improved considerably. The Neural ODE fix reduces the final mean position and velocity errors by $70\%$ for the transfer and by $98\%$ and $40\%$ for the landing. DaGGER reduces the final mean position and velocity errors by $14\%$ and $28\%$ for the transfer and by $22\%$ and $15\%$ for the landing.
We also refined G\&CNETs on a single trajectory using the Neural ODE fix. In the case of the interplanetary transfer we improved the final position error from  $1,241,662$ km to $2991$ km ($99\%$ reduction) and final velocity error from $9\cdot10^{-2}$ km/s to $5\cdot10^{-4}$ km/s ($99\%$ reduction). For the asteroid landing problem we improved the final position error from  $452$ m to $5.4$ m ($99\%$ reduction) and final velocity error from $0.9$ m/s to $0.5$ m/s ($44\%$ reduction). 
We found that the main advantages of DaGGER are twofold. First, it allows us to collect states, with their corresponding optimal controls, which are likely to be encountered by the G\&CNET, thereby forming a diverse dataset. Second, the computational effort required scales well with the size of the neural network, in contrast to the Neural ODE fix which is better suited for small networks due to the large amount of ODE sensitivities which need to be computed during each iteration. The downsides of DaGGER are that one constantly needs to query an expert, in our case this involves solving TPBVPs with a shooting method. Hence, one constantly runs the risk of injecting suboptimal trajectories (local minima) in the training dataset. In addition, for difficult problems such as the mass-optimal landing where a continuation (homotopy) approach is required to even find a solution, the DaGGER approach suffers from long convergence times or sometimes no convergence at all. Finally, the loss function and training hyper-parameters need to be carefully chosen for DaGGER such as to avoid catastrophic forgetting. The main advantages of the Neural ODE fix are that it allows us to learn from the dynamics and it is possible to give some final state errors more weight than others, thereby correcting what is more important to the user. Finally, in our experiments the Neural ODE fix did not cause catastrophic forgetting.
\section{Conclusion}
The use of Neural ODEs to improve the accuracy of Guidance \& Control Networks has been studied.
Both a time-optimal interplanetary transfer and a mass-optimal asteroid landing are considered. In all cases, we find that the final position and velocity errors to the target can be substantially reduced, both for a single trajectory and for a batch of 500 different trajectories. 
In the case of the interplanetary transfer the final position and velocity errors were reduced by $99\%$ for a single trajectory and the mean position and velocity errors were reduced by $70\%$ for a batch of trajectories.
In the case of the asteroid landing the final position and velocity errors were reduced by $99\%$ and $44\%$ for a single trajectory and by the mean position and velocity errors were reduced by $98\%$ and $40\%$ for a batch of trajectories. 
We were not able to reduce the final errors as drastically with the popular DaGGER approach. Nevertheless we acknowledge the strength and weaknesses of both approaches, most notably the fact that the Neural ODE fix is only suited for small networks, due to the computational burden associated with computing the ODE sensitivities at each iteration.
\vspace{-0.3cm}
\section*{Appendix}
\appendix
\section{Optimal control problems values}
\label{app:1}
\begin{table*}[!ht]
    \centering
    \caption{Initial conditions, final conditions and constants used for optimal control problems. \label{tab:val1}}
    \begin{tabular}{c|cc|cc|c||cc}
    
Problem & \multicolumn{2}{|c|}{Initial conditions}  & \multicolumn{2}{|c|}{Final conditions}  & Frame & \multicolumn{2}{|c}{Constants} 

\\\hline \hline

\multirow{5}{*}{Interplanetary} 
& $x_0$ & $-1.1874388$ AU & $x_f$ ($R$) & $1.3$ AU & \multirow{6}{*}{$\mathcal F$} &  $\Gamma$ & $0.1$ mm/s$^2$ \\ 
\multirow{5}{*}{transfer} & $y_0$ & $-3.0578396$ AU & $y_f$ & $0$ AU &  & $\mu$    &  (Sun) m$^3$/s$^2$\\ 
& $z_0$ & $0.3569406$ AU & $z_f$ & $0$ AU && \\ 
& $v_{x_0}$ & $-48.17$ km/s & $v_{x_f}$ & $0$ km/s && \\ 
& $v_{y_0}$ &  $18.30$ km/s & $v_{y_f}$ & $0$ km/s &&  \\ 
& $v_{z_0}$ &  $0.64$ km/s & $v_{z_f}$ & $0$ km/s  && \\ 

 \hline\hline

\multirow{6}{*}{Asteroid} 
& $x_0$ & $100$ km & $x_f$ & $30.27$ km & \multirow{6}{*}{$\mathcal R$} &    $I_{sp}$   &  $600$ s     \\ 
\multirow{6}{*}{landing}& $y_0$ & $150$ km & $y_f$ & $90.33$ km & &  $g_0$   &   $9.8$ m/s$^2$  \\ 
& $z_0$ & $0$ km & $z_f$ & $32.09$ km  & &   $c_1$   &  $80$ N    \\ 
& $v_{x_0}$ & $0.025$ km/s & $v_{x_f}$ & $0$ km/s & &  $\mu$    &    $1530348199$ m$^3$/s$^2$    \\ 
& $v_{y_0}$ &  $-0.025$ km/s & $v_{y_f}$ & $0$ km/s & &   $\omega$  &  $0.00041596$ rad/s   \\ 
& $v_{z_0}$ &  $0.02$ km/s & $v_{z_f}$ & $0$ km/s  &&      \\
\cline{2-6}
& $m_0$ &  $600$ kg & $m_f$ & left free  & -   &&    \\

 \hline\hline
    \end{tabular}
\end{table*}
    






\bibliographystyle{ISSFD_v01}
\bibliography{references}

\end{document}